\documentclass{article}
\PassOptionsToPackage{square, numbers}{natbib} 
\usepackage[preprint]{neurips_2024}

\usepackage[utf8]{inputenc} 
\usepackage[T1]{fontenc}    
\usepackage{hyperref}       
\usepackage{url}            
\usepackage{booktabs}       
\usepackage{amsfonts}       
\usepackage{nicefrac}       
\usepackage{microtype}      
\usepackage{xcolor}         
\usepackage[ruled,vlined]{algorithm2e}
\usepackage{graphicx}
\usepackage{soul}
\usepackage{tikz}
\usepackage{amssymb}
\usepackage{lipsum}
\usepackage{wrapfig}
\usepackage{caption}
\usepackage{mathtools}
\usepackage{multirow}
\usepackage{algpseudocode}
\usepackage{comment}
\usepackage{subcaption}
\usepackage{amsmath}
\usepackage{enumitem}
\setlist[enumerate]{itemsep=0mm}
\setlist{nosep}

\newcommand{\turbo}{{\texttt{GPT3.5Turbo}}\xspace}

\newcommand{\gptt}{{\texttt{GPT-4Turbo}}\xspace}
\newcommand{\mix}{{\texttt{Mixtral}}\xspace}
\newcommand{\indic}{{\texttt{IndicQA}}\xspace}
\newcommand{\tydi}{{\texttt{TyDiQA}}\xspace}
\newcommand{\squad}{{\texttt{SQuAD-F1}}\xspace}
\newcommand{\meta}{{\texttt{MLQA-F1}}\xspace}
\newcommand{\mono}{{\texttt{Mono}}\xspace}
\newcommand{\trans}{{\texttt{Trans}}\xspace}
\newcommand{\siml}{{\texttt{Sim}}\xspace}
\newcommand{\aggsrc}{{\texttt{Agg\_Src}}\xspace}
\newcommand{\aggtrans}{{\texttt{Agg\_Trans}}\xspace}
\newcommand{\gpteval}{{\texttt{GPTAnnotator}}\xspace}
\newcommand{\gptevalf}{{\texttt{GPTAnnotator-F1}}\xspace}
\newcommand{\haf}{{\texttt{HumanAnnotator-Score}}\xspace}

\title{Bridging the Gap: Dynamic Learning Strategies for Improving Multilingual Performance in LLMs}

\newcommand*\samethanks[1][\value{footnote}]{\footnotemark[#1]}

\author{%
Somnath Kumar\thanks{Equal Contributions} \quad Vaibhav Balloli\samethanks \quad Mercy Ranjit \quad Kabir Ahuja \quad Tanuja Ganu \\ \quad \textbf{Sunayana Sitaram} \quad \textbf{Kalika Bali} \quad \textbf{Akshay Nambi} \\
Microsoft Research India \\
\texttt{\{akshayn, taganu\}@microsoft.com}\\
}

\begin{document}

\maketitle

\begin{abstract}
Large language models (LLMs) are at the forefront of transforming numerous domains globally. However, their inclusivity and effectiveness remain limited for non-Latin scripts and low-resource languages. This paper tackles the imperative challenge of enhancing the multilingual performance of LLMs without extensive training or fine-tuning. Through systematic investigation and evaluation of diverse languages using popular question-answering (QA) datasets, we present novel techniques that unlock the true potential of LLMs in a polyglot landscape. Our approach encompasses three key strategies that yield significant improvements in multilingual proficiency. First, by meticulously optimizing prompts tailored for polyglot LLMs, we unlock their latent capabilities, resulting in substantial performance boosts across languages. Second, we introduce a new hybrid approach that synergizes LLM Retrieval Augmented Generation (RAG) with multilingual embeddings and achieves improved multilingual task performance. Finally, we introduce a novel learning approach that dynamically selects the optimal prompt strategy, LLM model, and embedding model per query at run-time. This dynamic adaptation maximizes the efficacy of LLMs across languages, outperforming best static and random strategies. Additionally, our approach adapts configurations in both offline and online settings, and can seamlessly adapt to new languages and datasets, leading to substantial advancements in multilingual understanding and generation across diverse languages.
\end{abstract}

\section{Introduction}
\label{sec:introduction}
Large Language Models (LLMs) such as ChatGPT~\cite{openai2023gpt4}, Gemini~\cite{team2023gemini}, and Claude~\cite{anthropic2023claude3} have revolutionized AI research, demonstrating vast advancements across various tasks~\cite{DBLP:journals/corr/abs-2005-14165,ouyang2022training,openai2023gpt4}. These models serve as intelligent assistants in enterprise applications (e.g., search engines, office suites), and several practical domains like healthcare, education, and agriculture~\cite{shiksha, farmerchat, khanmigo, m365copilot}. They are transforming teaching methods, agricultural practices, and many professional interactions with AI systems.

\vspace{-5pt}
\begin{wraptable}{r}{5cm}
\resizebox{0.3\textwidth}{!}{%
\begin{tabular}{ll}
\hline
Method              & Accuracy                    \\ \hline
LLama2 70B          & {8.5}  \\
Mistral 7B instruct & {29.6} \\ \hline
Cohere              & {78.8} \\
Palm2               & {76.5} \\
GPT3.5              & 60.1                        \\
GPT4                & 71.5                        \\ \hline
TULR-XXL            & \textbf{84.6}      \\ \hline        
\end{tabular}%
}\vspace{-5pt}
\caption{Performance comparison across various models for \tydi.}
\label{tab:motiv}
\vspace{-10pt}
\end{wraptable}
However, the current landscape primarily favors LLMs optimized for English and Latin script languages, limiting their effectiveness in non-English contexts~\cite{mega,ahuja2023megaverse,khanuja2021muril}. Despite recent advancements like fine-tuned LLMs~\cite{gala2024airavata} and smaller language models~\cite{abdin2024phi}, their performance remains subpar, especially in multilingual scenarios. Numerous studies highlight a noticeable performance gap between LLMs (including proprietary models like GPTx and Gemini, fine-tuned variants, and smaller models like OpenHathi~\cite{sarvamai2024openhathi}) and state-of-the-art (SOTA) multilingual models such as TULRv6 and XLMR~\cite{goyal2021larger}. 
Table~\ref{tab:motiv} compares the performance of various LLMs, including LLama2 70B, Mistral 7B instruct, Cohere, Palm2, GPT3.5, and GPT4, against SOTA models like TULR-XXL on the multilingual QA dataset \tydi (covering 9 diverse languages)~\cite{tydiqa}. Although GPT4 and Palm2 have improved over GPT3.5, a significant gap remains to SOTA models. This pattern persists across other multilingual QA benchmarks, such as MLQA, IndicQA, and AfriQA\cite{ogundepo2023afriqa}.

To bridge this gap, two main approaches are being explored. The first involves enhancing foundational model training, which faces several challenges: \textbf{1) Scarcity of Training Data:} Limited high-quality data for non-English and low-resource languages hinders optimal multilingual LLM pre-training~\cite{hämmerl2022multilingual, DBLP:journals/corr/abs-2004-06748}. \textbf{2) Lack of Access and Resources:} Many existing models are not open-source, and the high computational costs of training/fine-tuning restrict customization for specific languages~\cite{qin2024multilingual,liu2024understanding}. \textbf{3) Limited Adaptability:} Fine-tuned or smaller models often struggle to understand languages outside the targeted subset, despite performance improvements for specific languages~\cite{sitaram-etal-2023-everything}.

The alternative approach enhances the performance of pre-trained LLMs through external configurations: \textbf{1) Prompt Tuning:} Research explores strategies like native language, cross-lingual, and Chain-of-Thought prompting, improving performance for specific languages/tasks~\cite{weichain,shi2022language}. However, no single strategy consistently outperforms others across all tasks and languages~\cite{liu2024translation}. \textbf{2) Optimizing Model Embeddings:} Tasks like question answering benefit from Retrieval Augmented Generation (RAG), incorporating external knowledge~\cite{gao2023retrieval}. Improved text-embedding models enhance performance by retrieving relevant information. For example, OpenAI's text-embedding-3 improves multilingual performance over its predecessor, text-embedding-ada-002 (MIRACL score from 31.4\% to 54.9\%\cite{openai2024ada3}), and Cohere's embed-multilingual-v3.0 achieves a MIRACL of 67\%\cite{cohere2024embedv3}. Choosing the right embedding model remains challenging. \textbf{3) Model Selection Dilemma:} New LLM versions, proprietary (e.g., OpenAI GPTx, Gemini, Claude) and open-source (e.g., Llama, OpenHaithi), are continuously released. Identifying the best model for specific tasks and languages remains uncertain.

The lack of a definitive configuration (prompt strategy, embeddings, LLM model) drives our work, aiming to enhance multilingual LLMs performance without training or fine-tuning. We introduce three techniques to improve multilingual LLM performance: (i) optimizing prompting strategies, (ii) integrating multilingual embeddings with LLMs for better document retrieval and generation, and (iii) dynamically selecting prompts, language models, and embeddings at runtime to optimize performance across languages and tasks. Our key contributions are: 
\begin{enumerate}
\itemsep0em 
    \item \textbf{Optimizing Prompts for Polyglot LLMs:} Crafting prompts tailored to the unique characteristics of LLMs results in significant performance enhancements across languages.
    \item \textbf{Hybrid Approach combining LLM Generation with Multilingual Embeddings:} Combining LLM response generation with \emph{multilingual embeddings} in a Retrieval Augmented Generation (RAG) setting enhances the coherence and context relevance in text retrieval and generation, thereby enhancing multilingual task performance.
    \item \textbf{Dynamic Learning Approach for Performance Optimization:} Our approach dynamically selects the best prompt strategy, LLM model, and multilingual embedding model at runtime, maximizing efficacy across languages and surpassing best static and random strategies.
\end{enumerate}

We assessed the effectiveness of our techniques on two widely used Question Answering (QA) datasets: \indic and \tydi, covering 18 languages. Our findings reveal significant limitations in current datasets and evaluation approach, highlighting the need for improved task evaluation (see Section~\ref{sec:limitations}). Through extensive experimentation, \textbf{our results demonstrate over 15-20\% improvement in multilingual performance across diverse languages}. These results underscore the effectiveness of our approach, which can be applied to various prompting techniques, tasks, and models described in existing literature. We also provide the source code for the community\footnote{The source code will be released soon.}.

\vspace{-15pt}
\section{Multilingual Tasks, Datasets \& their Limitations}
\label{sec:setup}
\vspace{-10pt}
In this work, we prioritize Question Answering (QA) tasks, showcasing the model's ability to deliver accurate responses. Here, we focus on the RAG based QA tasks that leverage the text information provided as external knowledge-base for answering the question.
\vspace{-12pt}
\subsection{Dataset}
\label{sec:dataset}
\vspace{-5pt}
We utilize two prominent multilingual QA datasets:

1.\indic~\cite{ai4bharat2022indicqa}: A curated dataset in 11 Indic languages sourced from Wikipedia on topics related to Indic culture and history, comprising over 18,000 questions. 

2.\tydi~\cite{tydiqa}: This dataset covers 9 typologically diverse languages and includes two sub-tasks: passage selection and minimum answer span (Gold-P). Our experiments focus on the Gold-P task, where only the gold answer passage is provided rather than the entire Wikipedia article. 
\vspace{-5pt}
\subsection{Evaluation Metrics for Multilingual QA Task}
\label{subsec:eval_metrics}
\vspace{-5pt}
F1 score is the commonly used metric in QA tasks~\cite{DBLP:journals/corr/RajpurkarZLL16}, compares individual words in predictions to the True Answer. While \squad is standard for English QA evaluation, \meta~\cite{DBLP:journals/corr/abs-1910-07475} offers additional preprocessing for fair multilingual evaluation, including stripping Unicode punctuations and stand-alone articles. Hence, we adopt \meta as our evaluation metric. 

\begin{figure*}[!t]
\vspace{-5pt}
\begin{minipage}[l]{0.5\linewidth}
        \includegraphics[width=0.95\columnwidth]{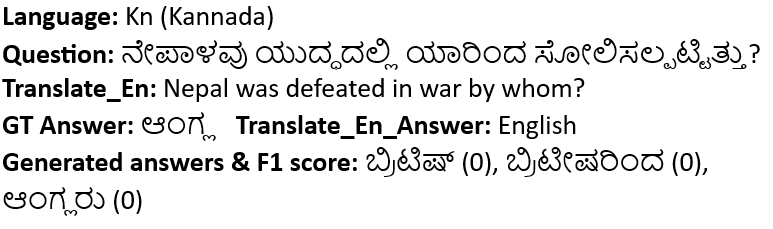}
\end{minipage}
\begin{minipage}[l]{0.5\linewidth}
       \includegraphics[width=0.95\columnwidth]{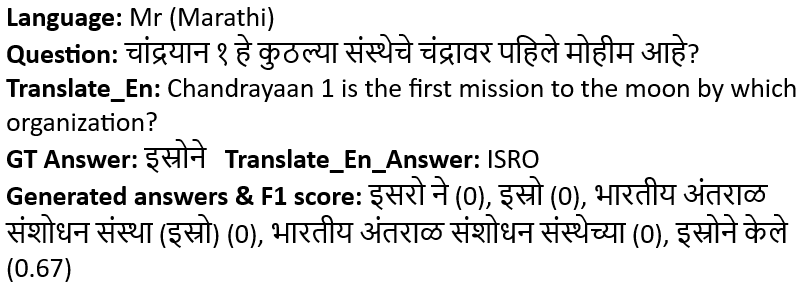}
\end{minipage}
\vspace{-5pt}
 \caption{Examples showing the limitations in the GT answer in \indic dataset.}
        \label{fig:examples}
        \vspace{-18pt}
\end{figure*}
\vspace{-5pt}
\subsection{Limitations of Current Datasets \& Evaluation Approach}
\label{sec:limitations}
Many public datasets for evaluating multilingual performance were developed before the Large Language Model (LLM) era. Evaluating LLMs on these datasets poses two main challenges:

\texttt{Challenge 1: Limited Ground Truth (GT).} GT in these datasets typically includes only one answer per question, whereas in reality, there could be multiple semantically equivalent variants more suitable in conversational or real-world application settings.\\
\texttt{Challenge 2: Strict Evaluation Approach.} The F1 score, often computed at the individual word level, poses a challenge especially in cases where there is only one GT. Even minor differences between ground truth and predicted answers lead to notable score reductions.

Figure~\ref{fig:examples} illustrates the above challenges on the \indic dataset for Kannada (kn) and Marathi (mr) languages. Each question is accompanied by the GT answer and generated results from different strategies. In the left example for Kannada, although the generated answers differ, they remain factually accurate, providing an alternative reference to "British". Similarly, in the right, the answer is an abbreviation for an organization, while the generated answers include a mix of abbreviated and fully expanded versions. As the dataset includes only one GT answer, the \meta scores for these cases would be exceptionally low, even zero. These examples highlight the limitations of current datasets and evaluation approach.

One potential solution is to enrich the ground truth with all possible alternatives. However, this would require significant data collection effort, proving both expensive and cumbersome.
Recently, LLMs have proven to be proficient annotators for various tasks across domains~\cite{he2023annollm}. Verifying or validating answers is simpler than generating them~\cite{kuchnik2023validating}. To address the GT limitation, we propose \gpteval, utilizing an LLM to validate a set of predicted answers. \gpteval employs an LLM like GPT4, where for each question, it assesses numerous predicted answers generated using different strategies (prompts and models) for their correctness (see Appendix~\ref{sec:gptanno} for prompts).

\gpteval offers three options: \textit{YES} for any semantically correct answer that matches with the GT answer, \textit{NO} for no match, and \textit{PARTIAL} for partial match. When \gpteval yields a \textit{YES}, predicted answers are added to the original GT, enhancing it with multiple equivalent answers. For instance in Figure~\ref{fig:examples}, all generated answers would now be marked as correct and added to GT.
This addresses challenge 1 by ensuring a more comprehensive evaluation. To tackle challenge 2, we introduce \gptevalf score, computing F1 against this new comprehensive ground truth with multiple semantically correct answers. Thus, we evaluate LLMs' multilingual performance for QA tasks using two metrics: \meta, comparing the predicted answer to the original GT, and \gptevalf, evaluating against a derived, comprehensive GT. Note both are F1 scores but evaluated with single answer GT (\meta) and multiple answers GT (\gptevalf).

Our goal is to ensure that the \gptevalf score closely mirrors a human annotator's score. To compare \gptevalf and \meta, we randomly selected 100 questions from the \indic dataset across six languages (bn, hi, ta, kn, mr, gu) and presented them to native speakers in our research group. Each annotator received the context paragraph, question, GT answer, and a set of predicted answers generated using various strategies, akin to \gpteval. Annotators selected \textit{YES} for a complete match (Score = 1), \textit{NO} for no match (Score = 0), and \textit{PARTIAL} for a partial match (Score = 0.5 or F1 against GT), termed \haf.

\begin{wrapfigure}{r}{5cm}
\vspace{-8pt}
    \centering
\includegraphics[width=0.4\columnwidth, height=1.2in]{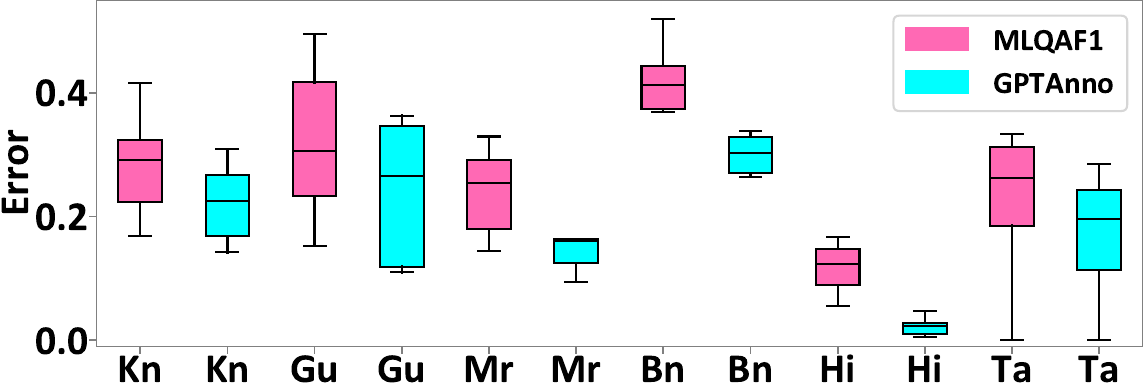}
        \caption{Comparison of \meta and \gptevalf. }
        \label{fig:error_metrics_ha}
        \vspace{-12pt}
\end{wrapfigure}
Figure~\ref{fig:error_metrics_ha} illustrates the difference between \haf and \meta, and \haf and \gptevalf for the six languages across all strategies. On these 100 samples (across 6 languages), \meta scores showed an average difference of 25\% and a maximum difference of 51\% compared with \haf, indicating significant gap. This discrepancy arises due to the limited original GT provided in the datasets and the generative models' ability to produce responses with variations. In contrast, \gptevalf differences are substantially lower compared to \meta differences when compared to \haf. Additionally, across all languages, \gptevalf scores reduce the difference by 30\% compared to \meta scores and align more closely with \haf. Thus, with the assistance of \gpteval, we can address current limitations in the dataset and evaluation approach, and importantly, evaluate performance more akin to human assessment. In the subsequent sections, we present results obtained using both \meta and \gptevalf metrics.

\vspace{-15pt}
\section{Prompt Strategies for Polyglot LLMs}
\label{sec:prompts}
\vspace{-5pt}
Generative models' performance heavily relies on prompt tuning~\cite{sahoo2024systematic}. Crafting effective prompts, even for specific English tasks, is challenging~\cite{yang2022prompt}. Although tools like PromptBreeder~\cite{fernando2023promptbreeder} and MedPrompt~\cite{chen2023medprompt} aid prompt creation for generic English QA tasks, defining prompts for multilingual scenarios lacks clear strategies~\cite{mega,ahuja2023megaverse}. Through extensive experiments and analysis, we've identified LLM strengths and designed optimized templates for polyglot LLMs.

Our notation is as follows: Each prompt comprises an Instruction and Context, with contexts being passages like in reading comprehension for answering questions in QA scenarios. We employ two approaches - \textit{zero-shot}, where no examples are passed, and \textit{few-shot}, where a few random examples are passed based on the strategy. The language in which the query is issued is termed the source language, which is also the language for the final response. We devise the following prompting strategies to enhance the model's performance in multilingual tasks:

\textbf{1. Monolingual (Mono):}
In this technique, the Instruction and Context are in the source language. Few-shot examples are also from the source language.

\textbf{2. Translate-Test (Trans)}: 
This approach translates the Instruction and Context into English using an automatic Machine Translation system~\cite{azuretranslate}. The model is then queried in English, and the result is back-translated into the source language (Roundtripping through English).

\textbf{3. Similar high-resourced language (Sim):} Similar to translate-test, this method involves roundtripping through another pivot language, typically in the high or medium-resource category. The pivot language is selected based on its proximity to the source language in terms of language feature similarities~\cite{malaviya17emnlp} captured in lang2vec~\cite{littell2017uriel}. The rationale is that selecting a pivot language based on language similarities allows for better capture of linguistic aspects compared to direct translation to English. Initially, multiple similar languages are identified using various feature similarity metrics, from which the pivot language is chosen. Preference is given to high-resource languages with Latin script, as evidence suggests they perform better~\cite{liu2024translation}. This selection process ensures the pivot language, similar to the source language, enhances the strategy's effectiveness. More details in Appendix~\ref{app:sim}.

\textbf{4. Aggregation Source (Agg\_Src):} This strategy aggregates responses from previous strategies such as mono, trans, and sim. These responses are presented to the LLM to determine the best answer in the source language based on input from other strategies. While this increases the number of calls to the LLM, it effectively combines the strengths of different prompting strategies and leverages information from multiple languages to generate a single response.

\textbf{5. Aggregation  Translate (Agg\_Trans)} Similar to Agg\_Src, this strategy collects responses from each of the other strategies, such as mono, trans, and sim. These responses are translated into English. The three English responses are then aggregated using the LLM to generate a single aggregated response in English. This aggregated response in English is subsequently translated back into the source language to obtain the final response.

While we emphasize the above five primary multilingual prompting strategies, we also experimented with other popular approaches like self-translation and Chain-Of-Thought, which showed inconsistent performance. All strategies are executed with both zero-shot and few-shot (in-context examples). We use randomly selected fixed few-shot examples for consistency, aiming to demonstrate performance improvement. Advancements in in-context learning and example selection will further enhance all strategies and are complementary.

\begin{table}
\centering
\resizebox{0.8\textwidth}{!}{%
\begin{tabular}{llll|lll}
\hline
 & \multicolumn{3}{c|}{\meta}                          & \multicolumn{2}{c}{\gptevalf} &                     \\ \hline
 & \gptt & \turbo & \mix & \gptt & \turbo & \mix \\ \hline
\mono     & \textbf{0.51} & \textbf{0.43} & 0.15 & 0.71 & 0.71 & 0.31 \\
\trans     & 0.36 & 0.37 & \textbf{0.33} & \textbf{0.80} & \textbf{0.80} & \textbf{0.68} \\
\siml     & 0.30 & 0.28 & 0.19 & 0.70 & 0.70 & 0.44 \\
\aggsrc   & \textbf{0.51} & \textbf{0.43} & 0.20 & 0.73 & 0.73 & 0.39 \\
\aggtrans & 0.35 & 0.38 & \textbf{0.33} & 0.79 & 0.79 & \textbf{0.68} \\ \hline
\end{tabular}%
}
\caption{Performance of different Prompting strategies for \indic. Bold indicates the strategies that give best performance for that model.}
\label{tab:prompt_indic}
\vspace{-20pt}
\end{table}

\textbf{Prompting Strategies Results.}
We found that few-shot examples consistently improve performance across all models compared to zero-shot. Therefore, we present only few-shot results going forward. Table~\ref{tab:prompt_indic} shows average performance of the individual prompting strategies across all languages in the \indic dataset across \gptt~\cite{azurepoenaigptfour}, \turbo~\cite{azureopenaiturbo}, and \mix~\cite{azureopenai}, considering both \meta and \gptevalf metrics with text-embedding-ada-002 default embeddings.  Three main takeaways emerge:\\
\texttt{1. No Universal Best Strategy:} There is no single prompt strategy that universally performs best across all languages for each model. For \turbo and \gptt, \mono and \aggsrc strategies generally excel across most languages, while \mix tends to favor \trans and \aggtrans. Notably, languages like `ta' and `te' consistently prefer translate strategies due to low-resource language constraints. Thus, the optimal strategy varies for each model and language. Appendix~\ref{app:prompt} provides more details regarding per-language performance across prompt strategies.

\texttt{2. Strategy Sensitivity to Metrics:}  The choice of prompt strategies changes when different metrics are considered. When \gptevalf scores are utilized for \gptt and \turbo, the best strategies for all languages include \trans and \aggtrans, whereas \mono and \aggsrc perform better when \meta is used. Similar takeaways arise for the \tydi dataset see Appendix~\ref{app:prompt}.


\textbf{Summary: }\textit{Prompt strategies enhance multilingual performance, but there's no one-size-fits-all solution across datasets, metrics, models, and languages. Notably, with \gptevalf scores, \turbo shows comparable performance to \gptt.} 

\vspace{-10pt}
\section{Hybrid Approach: Synthesizing LLM Generation with Multilingual Embeddings}
\label{sec:hybrid}
Currently, most LLMs are trained predominantly on English and a few high-resource languages, limiting their understanding of medium and low-resource languages~\cite{huang2023not}. Tasks like question answering and summarization often require LLMs to incorporate non-parametric knowledge from private knowledge bases or contextual information. Retrieval Augmented Generation (RAG) addresses these limitations by retrieving relevant documents or information chunks for a given query and then synthesizing responses~\cite{gao2023retrieval}. The process involves: 1) encoding knowledge-base documents using a text embedding model and indexing them for retrieval speed, 2) encoding the query using the same model, 3) conducting a similarity search between the query embedding and the indexed document embeddings to retrieve top $k$ similar documents, and 4) forwarding the retrieved documents to the LLM for response synthesis. Better text-embedding models enhance task performance, especially in multilingual scenarios, by fetching the most relevant information based on the query. This improves response generation, even in languages with limited training data.

While LLMs excel in response synthesis, enhancing multilingual task performance relies on a robust multilingual text-embedding model. Since GPT models are primarily trained on English data, their default embedding model (text-embedding-ada-002, or ada) leads to suboptimal performance. In contrast, state-of-the-art multilingual models like XLMR-XXL~\cite{goyal2021larger}, and Cohere (embed-multilingual-v3.0)~\cite{cohere2024embedv3} demonstrate superior performance due to their training on diverse language data.

In this paper, we introduce a \textbf{hybrid approach} that integrates the cross-lingual semantic understanding of multilingual embeddings with the text generation capabilities of LLMs. We experiment with GPT's default embedding (ada), its improved variant, text-embedding-3-large (adav3)~\cite{openai2024ada3}, which offers enhanced multilingual performance, and state-of-the-art multilingual embeddings such as XLMR-XXL~\cite{goyal2021larger} and Cohere multilingual embeddings v3 (Cohere)~\cite{cohere2024embedv3}.

\textbf{Performance Analysis.} Table~\ref{tab:hybrid-perf} illustrates the maximum performance achieved by each embedding (ada, adav3, xlmr, cohere) for \gptt and \turbo models across all languages and prompt strategies for \indic. Cohere, a multilingual embedding, enhances \gptt performance by up to 7\% and 2\% compared to default ada embeddings when using \meta and \gptevalf metrics. This indicates a significant improvement in multilingual task performance with multilingual embeddings coupled with LLM generation. While marginal improvements are observed in \turbo with multilingual embeddings, mainly due to poor LLM generation with \turbo rather than multilingual content retrieval. 

Additionally, Table~\ref{tab:emd-pre-indic} and~\ref{tab:emd-pre-tydi} indicates the preferred embedding for each language that yields the best performance. Generally, multilingual embeddings, particularly Cohere, are preferred for \indic. Similar trends are observed in \tydi, as detailed in Appendix~\ref{app:hybrid}. 
 
\textbf{Summary:} \textit{The hybrid approach incorporating multilingual embeddings enhances task performance by up to 7\% on the \gptt model. However, there's no universal best prompt strategy, model, or embedding that performs optimally across datasets and languages.}

\begin{table*}[!t]
\begin{minipage}[l]{0.33\linewidth}
\renewcommand{\arraystretch}{3.0}
\resizebox{\columnwidth}{!}{%
\begin{tabular}{llllll}
\hline
Metrics                                   & Models                & Ada  & Adav3 & XLMR & Cohere \\ \hline
\multirow{2}{*}{\meta}     & \gptt  & 0.51 & 0.5   & 0.54 & \textbf{0.58}   \\
                                          & \turbo & 0.43 & 0.43  & 0.39 & \textbf{0.44}   \\ \hline
\multirow{2}{*}{GPTAnno} & \gptt  & 0.8  & 0.8   & 0.8  & \textbf{0.82}   \\
                                          & \turbo & 0.8  & 0.8   & 0.8  & \textbf{0.81}   \\ \hline
\end{tabular}%
}
\caption{Hybrid approach performance on \indic.}
\label{tab:hybrid-perf}
        \vspace{-5pt}
\end{minipage}
~~
\begin{minipage}[l]{0.3\linewidth}
\renewcommand{\arraystretch}{1.5}
\resizebox{1.0\textwidth}{!}{%
\begin{tabular}{l|ll|ll}
\hline
     & \multicolumn{2}{l|}{\meta}    & \multicolumn{2}{l}{\gptevalf} \\ \hline
Lang & \turbo & \gptt & \turbo  & \gptt \\ \hline
as & Ada    & Cohere & Ada    & Ada    \\
bn & Cohere & Cohere & Ada    & Ada    \\
gu & Ada    & Cohere & Cohere & Cohere \\
hi & Cohere & Cohere & Ada    & Cohere \\
kn & Ada    & Cohere & Cohere & Cohere \\
ml & Cohere & Cohere & Cohere & Cohere \\
mr & Ada    & Cohere & Cohere & Cohere \\
or & Adav3   & Ada    & Ada    & Ada    \\
pa & Adav3   & Adav3   & Cohere & Ada    \\
ta & Cohere & Cohere & Cohere & Cohere \\
te & Cohere & Cohere & Cohere & Cohere \\ \hline
\end{tabular}%
}
\caption{Embedding preference \indic.}
\label{tab:emd-pre-indic}
        \vspace{-5pt}
\end{minipage}
~~
\begin{minipage}[l]{0.31\linewidth}
\renewcommand{\arraystretch}{1.5}
\resizebox{1.0\textwidth}{!}{%
\begin{tabular}{l|ll|ll}
\hline
 & \multicolumn{2}{l|}{\meta}    & \multicolumn{2}{l}{\gptevalf} \\ \hline
Lang & \turbo & \gptt & \turbo  & \gptt \\ \hline
ar & Ada    & Adav3    & Ada    & Adav3    \\
bn & Ada    & Ada     & Ada    & Ada     \\
en & Cohere & XLMR & Cohere & XLMR \\
fi & Ada    & Ada     & Ada    & Ada     \\
id & Ada    & Adav3    & Ada    & Adav3    \\
ko & Ada    & XLMR & Ada    & XLMR \\
ru & Ada    & Ada     & Ada    & Ada     \\
sw & Ada    & Adav3    & Ada    & Adav3    \\
te & Ada    & Cohere  & Ada    & Cohere \\ \hline
\end{tabular}%
}
\caption{Embedding preference \tydi.}
\label{tab:emd-pre-tydi}
        \vspace{-5pt}
\end{minipage}
\vspace{-5pt}
        \vspace{-10pt}
\end{table*}
\vspace{-15pt}
\section{Learning Approach to Improve Multilingual Task Performance}
\label{sec:learning}
\vspace{-5pt}
As demonstrated in previous sections, it is evident that a one-size-fits-all approach does not exist when it comes to selecting the optimal configuration of prompt strategy, embeddings, and LLM model for diverse multilingual languages. This leads us to the crucial question: \textit{Can we dynamically determine the most optimal configuration for each query to maximize multilingual task performance?}

To tackle this challenge, we propose a learning approach that identifies the most optimal configuration for a given query. This approach should fulfill several key requirements: (i) \textbf{Offline Learning:} It should be capable of learning the optimal configuration using ground truth data in an offline setup, (ii) \textbf{Online Learning:} It should also be able to adapt and learn in real-time, allowing the base model to fine-tune itself to accommodate new data and distribution shifts between training and testing, and (iii) \textbf{Language and Dataset Adaptability:} Additionally, it should be flexible enough to adapt to different languages and datasets in an online setup, ensuring robust performance across diverse linguistic contexts and data sources. 

\textbf{Hybrid architecture.} Our proposed learning approach combines the capabilities of LLMs with convolutional layers to dynamically select the most optimal configuration per language and task. We utilize LLMs to process input data and generate high-dimensional representations. These representations form an ND array, with dimensions corresponding to LLM model, prompt strategy, and embeddings. Convolutional layers with ND operate on this array, extracting relevant features across dimensions and learning indicative representations for optimal configurations. Subsequently, the output is the prediction of task accuracy (F1 Score) for the particular query across all configurations. This hybrid architecture integrates LLMs and convolutional layers, achieving superior performance in configuration selection. By comparing predicted accuracies, we dynamically select the optimal configuration for each query, tailored to specific task and language requirements. This approach satisfies offline and online learning needs and adapts to various languages and datasets, ensuring effective configuration selection for diverse multilingual tasks.

Prior efforts like LOVM~\cite{zohar2024lovm}, \cite{liu2023towards}and HuggingGPT~\cite{shen2024hugginggpt} have mainly concentrated on optimizing the model selection for a task. These methods typically consider only a single parameter selection at a time. In contrast, our approach concurrently selects the optimal configuration across three parameters: LLM model, prompt strategy, and embeddings. This introduces a complex and high-dimensional search space, posing challenges beyond the scope of iterative processes in prior efforts.

In our hybrid architecture, which combines LLMs with Convolutional ND layers, we predict the F1 score for each configuration per query. These scores form the basis for generating a \emph{SoftMax} output, which serves as a probability density function. Sampling configurations from this distribution is crucial, especially in an online learning setting. This probabilistic sampling enables the control of entropy and encourages the exploration of diverse configurations, particularly with out-of-distribution data, thus helping to mitigate bias. 

\begin{figure}[t!]\centering
    \includegraphics[width=0.65\linewidth]{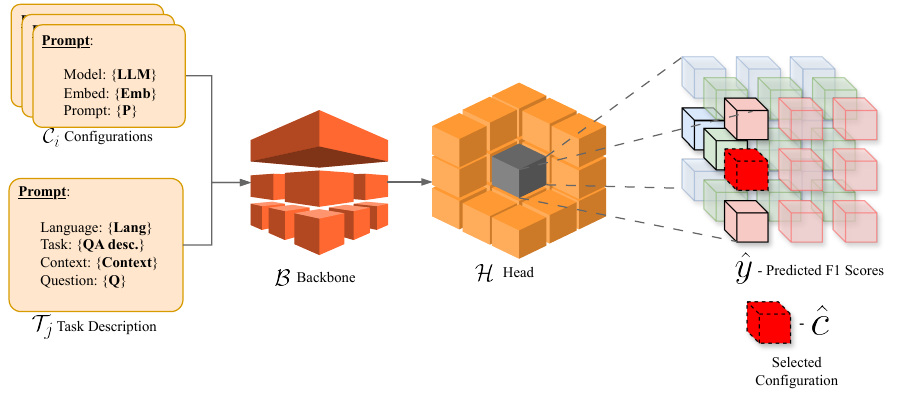}
    \vspace{-15pt}
    \caption{Illustration of Inference Pipeline}
    \label{app:fig:inference}
    \vspace{-15pt}
\end{figure}
\textbf{Architecture details.} The architecture utilizes LLaMa-2-70B-hf model to produce embeddings. The traditional sampling head is replaced by a set of Conv-ND layers \cite{vizcaino2021convnd}, denoted as \(\mathcal{H}\), which predict the F1 Score for each configuration. The LLaMa-2-70B-hf \cite{touvron2023llama} backbone, \(\mathcal{B}\), embeds the Task Description \(\mathcal{T}\), generating embeddings \(\mathcal{E}_T\). The backbone \(\mathcal{B}\) is also used to embed the individual configurations \(\mathcal{C}_i\) into embeddings \(\mathcal{E}_{C_i}\).

These embeddings are then arranged into an ND array of size \(\mathcal{R}^{e \times n_1 \times n_2 \times n_3 \dots n_m}\), where \(m\) is the number of parameters (e.g., language model, embedding model, prompt strategies, so \(m=3\)). Each \(n_i\) represents the number of possibilities for each parameter (e.g., three language models (\gptt, \turbo, \mix), four embedding models (adav2,adav3,XLMR,cohere), five prompt strategies (\mono, \trans, \siml, \aggsrc, \aggtrans)). The embedding projection size \(e\) for \(\mathcal{B}\) is 8192. The task embedding is broadcasted and concatenated to form a matrix of size \(\mathcal{R}^{2e \times n_1 \times n_2 \times n_3 \dots n_m}\). We treat the embedding dimension as the number of input channels to \(\mathcal{H}\) and reduce it to 1 while preserving the remaining dimensions, resulting in a matrix of size \(\mathcal{R}^{1 \times n_1 \times n_2 \times n_3 \dots n_m}\) or \(\mathcal{R}^{n_1 \times n_2 \times n_3 \dots n_m}\), representing the predicted F1 scores for all configurations.
\small
\begin{equation}
\mathcal{E}_{T_j} \leftarrow \mathcal{B}(\mathcal{T}_j);
\mathcal{E}_{C_i} \leftarrow \mathcal{B}(\mathcal{C}_i);
\mathcal{E}_j \leftarrow \mathcal{E}_{T_j} \parallel \mathcal{E}_{C_i} ;
\hat{y} \leftarrow \mathcal{H}(\mathcal{E}_j)
\end{equation}
\normalsize
Using the above, we obtain \(\hat{y}\), which is the predicted F1 score for all combinations. To select the configuration, we either take the \emph{argmax} or apply \emph{softmax} and sample a particular configuration. Figure. \ref{app:fig:inference} illustrates inference pipeline, given the Task Description \(\mathcal{T}_{j}\) and Configurations \(\mathcal{C}_{i}\) to obtain \(\hat{c}\) for sampled configuration.
\vspace{-5pt}
\subsection{Training the Model for Both Online and Offline Setups.} To train the Backbone \(\mathcal{B}\) and the Head \(\mathcal{H}\), we employ two different loss functions based on the specific scenario: offline and online settings. 

\textbf{1. Offline Setting:} In the offline setting, we have the advantage of knowing the F1 scores for all possible configurations for each given sample. This complete information allows us to obtain the ground truth F1 scores for all samples, denoted as \(y\). We can then use these ground truth F1 scores to train the backbone \(\mathcal{B}\) and the head \(\mathcal{H}\) effectively.
\begin{enumerate}
\itemsep0em 
 \item \textbf{Infer F1 Scores for All Configurations}: For each sample, infer the F1 scores for all possible configurations. This means computing the F1 scores for each combination of the language model, embedding model, and prompt strategies. For example, if there are three configurations for each parameter (e.g., three language models (\gptt, \turbo, \mix), four embedding models (adav2,adav3,XLMR,cohere), five prompt strategies (\mono,\trans, \siml,\aggsrc, \aggtrans)), we would infer F1 scores for \(3 \times 4 \times 5 = 60\) configurations per sample.
 \item  \textbf{Obtain Ground Truth F1 Scores}: Collect the actual F1 scores for all configurations, which serve as the ground truth \(y\). Thus, for each sample, we gather the F1 scores for all 60 configurations.
 \item  \textbf{Train Using MSE Loss}: Use the Mean Squared Error (MSE) loss to train the model. The MSE loss is computed between the predicted F1 scores \(\hat{y}\) and the ground truth F1 scores \(y\)  $   {MSE Loss} = \frac{1}{N} \sum_{i=1}^N (\hat{y}_i - y_i)^2     $,    where \(N\) is the number of samples.
\end{enumerate}

\textbf{Online Setting:} In the online setting, we only have the ground truth F1 score for the configuration that was selected and inferred. This results in a sparse matrix of F1 scores, as we do not compute the F1 scores for all configurations to avoid the computational cost. 
\begin{enumerate}
\itemsep0em 
\item \textbf{Infer F1 Score for Selected Configuration}: For each sample, infer the F1 score for only the selected configuration. This selected configuration is chosen based on the model's predictions or a sampling strategy. For example, if the model predicts or selects a specific configuration out of 60, we only compute the F1 score for that particular configuration.
\item  \textbf{Obtain Ground Truth F1 Score}: Compute the actual F1 score for the selected configuration, which serves as the ground truth \(y_{{selected}}\).
\item  \textbf{Update Using Sparse Matrix}: Update the model using the sparse matrix of predicted F1 scores \(\hat{y}\). Only the F1 score corresponding to the selected configuration is updated, leaving the other entries unaffected, thus reducing the computational overhead.
\item  \textbf{Adjust Loss Function}: The loss function must account for the sparsity. Instead of a straightforward MSE loss, we use a modified loss function that updates only the predicted F1 score for the selected configuration, 
 ${Sparse MSE Loss} = (\hat{y}_{{selected}} - y_{{selected}})^2 $, where \(\hat{y}_{{selected}}\) is the predicted F1 score for the selected configuration, and \(y_{{selected}}\) is the ground truth F1 score for the same configuration. This loss function ensures the model is updated based on the selected configuration without needing the complete F1 score matrix. Implementation details of the above pipeline is explained in Appendix~\ref{app:training}.
 \end{enumerate}

\textbf{Train-Test Split.} We split both datasets into three subsets, Offline training (60\%), Online adaptation (20\%), test set (20\%), respectively. Furthermore, we train and evaluate the models using the two evaluation approaches with metrics \meta and \gptevalf.

\begin{table*}[!t]
\begin{minipage}[l]{0.5\linewidth}
\renewcommand{\arraystretch}{2.2}
\resizebox{\textwidth}{!}{%
\begin{tabular}{|l|l|cc|cc|ccc|}
\hline
Evaluation & Datasets             & Acc & Acc & F1& F1       & Max           & Random & Best \\ 
& &@top1 & @top5& @top1&@top5 & F1& F1& single F1\\\hline
\multirow{2}{*}{\meta}     & \indic & 0.41 & 0.83 & 0.60 & \textbf{0.64} & 0.64          & 0.46 & 0.51 \\
           & \tydi & 0.57     & 0.78     & 0.52    & \textbf{0.54} & 0.54          & 0.43   & 0.50        \\ \hline
\multirow{2}{*}{GPTAnno} & \indic & 0.32 & 0.48 & 0.59 & 0.68          & \textbf{0.69} & 0.49 & 0.58 \\
           & \tydi & 0.62     & 0.55     & 0.56    & 0.69          & \textbf{0.72} & 0.51   & 0.54        \\ \hline
\end{tabular}%
}
\caption{Offline performance.}
\label{tab:offline}
        \vspace{-5pt}
\end{minipage}
\begin{minipage}[l]{0.5\linewidth}
\renewcommand{\arraystretch}{2.2}
\resizebox{\textwidth}{!}{%
\begin{tabular}{|l|l|cc|cc|ccc|}
\hline
Evaluation                            & Datasets              & Acc   & Acc   & F1    & F1            & Max           & Random & Best   \\
                                      &                       & @top1 & @top5 & @top1 & @top5         &  F1             &   F1     & single F1 \\ \hline
\multirow{2}{*}{\meta} & \indic & 0.29  & 0.73  & 0.60  & \textbf{0.63} & 0.63          & 0.46   & 0.51   \\
                                      & \tydi  & 0.62  & 0.85  & 0.51  & \textbf{0.52} & 0.52          & 0.41   & 0.45   \\ \hline
\multirow{2}{*}{GPTAnno}              & \indic & 0.52  & 0.57  & 0.62  & 0.66          & \textbf{0.69} & 0.52   & 0.61   \\
                                      & \tydi  & 0.62  & 0.67  & 0.73  & 0.74          & \textbf{0.76} & 0.54   & 0.69   \\ \hline
\end{tabular}
}
\caption{Online performance.}
\label{tab:online}
        \vspace{-5pt}
\end{minipage}
\vspace{-5pt}
        \vspace{-10pt}
\end{table*}

\vspace{-10pt}
\subsection{Evaluation of Learning Approach}
\vspace{-5pt}
\textbf{1. Offline Training Results.}
To evaluate our offline training phase, we compare against two baselines to gauge the effectiveness of our approach: (i) Randomly Selecting Configurations, representing sample variance, (ii) The Best Single Configuration, where the highest-scoring configuration across all samples is chosen. This ideal scenario reflects the performance when an optimal configuration is chosen as default.  We evaluate the performance using multiple metrics, Accuracy for selecting configurations, F1 score for task performance with the selected configurations, i.e., \textbf{Acc@Top1} (how often the correct configuration is predicted by our model), \textbf{Acc@Top5} (the accuracy of correct configuration in the top 5 predictions) and F1 score accuracy at top 1 and top 5 (what is the F1 score obtained by executing the configurations selected with top1 and top5).

Table~\ref{tab:offline} shows the performance of our model against random and best single configuration with both \meta and \gptevalf scores. Our model consistently outperforms random selection (by 17\%) and best single configuration (by 11\%) baselines for both scores (\meta and \gptevalf), indicating its ability to dynamically adapt and select configurations that optimize performance. Notably, our approach achieves a top 5 accuracy that matches the maximum achievable accuracy, underscoring its robustness in capturing diverse and potentially correct answers. 

\textbf{2. Online Training results:} In online training, we assessed our model's ability to adapt to new data distributions. We used the same baseline methods as in offline training and compared their performance with our approach on 20\% online adaptation set.
For online adaptation, we employed parameters ($\mathcal{B}$ and $\mathcal{H}$) from the offline training at epoch 100 and further trained the model for 10 epochs on the online adaptation set. The performance in the online test phase (Table \ref{tab:online}) showcases our model's remarkable adaptability, achieving F1 score accuracies at top 1 (60\%) and top 5 (63\%) closely matching the maximum achievable accuracy (63\%). This indicates its consistency in capturing optimal configurations across diverse datasets and languages. Furthermore, our approach significantly outperformed baseline methods, surpassing random selection by 15\% and the best single configuration by 7\%. Even with minimal fine-tuning epochs, our approach demonstrates its effectiveness in adapting to new or out-of-distribution data.


\begin{table}
\centering
\resizebox{\textwidth}{!}{%
\begin{tabular}{l|l|ll|ll|lll}
\hline
Evalaution & Languages & Acc@top1 & Acc@top5 & F1@top1 & F1@top5       & Max-
F1           & Random-F1 & Best Single-F1 \\ \hline
\multirow{3}{*}{\begin{tabular}[c]{@{}l@{}}Language\\ Adaptation\end{tabular}} &
  Kn &
  0.29 &
  0.75 &
  0.44 &
 \textbf{{0.46}} &
  {0.47} &
  0.37 &
  0.45 \\
           & Ta        & 0.28     & 0.74     & 0.48    & \textbf{{0.50}} & {0.53} & 0.43   & 0.46        \\
           & Te        & 0.28     & 0.74     & 0.51    & \textbf{{0.55}} & {0.57} & 0.43   & 0.49        \\ \hline
\begin{tabular}[c]{@{}l@{}}Dataset \\ Adaptation\end{tabular} &
  \begin{tabular}[c]{@{}l@{}}\tydi on \\ \indic base\end{tabular} &
  0.56 &
  0.67 &
  0.43 &
  \textbf{{0.52}} &
  {0.52} &
  0.41 &
  0.45 \\ \hline
\end{tabular}%
}
\caption{Learning approach performance on adaptation to unseen languages and datasets.}
\label{tab:adapt}
\vspace{-22pt}
\end{table}

\textbf{3. Adaptation Efficacy:} We now evaluate the adaptability of our approach.\\
\textit{(i) Adaptation to Unseen Languages:} To assess our model's adaptability to newer downstream languages, we conducted experiments on languages not encountered during offline training. This is essential for evaluating its ability to generalize across diverse languages, crucial for real-world applications. We trained the backbone and the head on the \indic dataset, excluding Kannada, Tamil, and Telugu languages. The excluded languages were then used for online training, simulating scenarios where the model encounters new languages during inference.
Our results, summarized in Table~\ref{tab:adapt}, demonstrate the model's remarkable adaptation capability.  It achieves performance levels comparable to maximum achievable scores across Kannada, Tamil, and Telugu, showcasing effective generalization across diverse linguistic contexts. Compared to single best and random baselines, our model consistently outperforms them in terms of F1 scores across all evaluated languages, indicating its effectiveness in adapting to new languages.\\
\textit{(ii) Adaptation to Different Datasets:} Adapting to diverse datasets with varying languages and query distributions is crucial for real-world applications. We trained the backbone and head of our model on the \indic dataset across all languages and tested on data from \tydi. Despite the small intersection of common languages between \indic and \tydi, our model demonstrated remarkable effectiveness, outperforming random selection by 11\% and the best single strategy by 7\% (see Table~\ref{tab:adapt}). With only 15 epochs of fine-tuning on 20\% of \tydi, we achieved the maximum F1 score, emphasizing the importance of adapting to diverse datasets, languages, and query distributions.


\textbf{Summary:} \textit{Our learning approach showcases significant improvements, underscoring its efficacy in dynamically selecting configurations, and adapting to new languages and datasets.}
\vspace{-12pt}
\section{Conclusions}
\vspace{-8pt}
In this work, we introduced a dynamic learning approach aimed at improving the performance of multilingual LLMs without requiring extensive training or fine-tuning. Our findings highlight several important insights. Firstly, we observed that prompting strategies lack universality, necessitating tailored approaches for different datasets, metrics, models, and languages. Additionally, our investigation into hybrid embeddings demonstrated performance enhancements, particularly with Cohere, which improved task performance by up to 7\% on the \gptt model. This underscores the significance of leveraging multilingual embeddings to augment LLM performance across varied linguistic contexts. Lastly, through offline and online assessments, our learning model surpassed baseline approaches by 15-20\% in dynamically selecting configurations and adapting to new data distributions. Overall, our dynamic learning approach presents a promising avenue for enhancing the multilingual capabilities of LLMs, with implications for various real-world applications. Future research directions include further exploration of learning techniques, scalability to larger datasets, and the generalization of our approach to other NLP tasks.\\
\textbf{Limitations and Broader Research:} 
While our work takes a first step towards improving multilingual performance, the system is still not fully inclusive, and as a community, we must explore ways to ensure LLMs are accessible to all. Finally, while our key contributions including learning algorithms are generalizable, the optimal strategies and embeddings may differ from one dataset to another. As the demand for multilingual language models continues to rise, our findings lay the foundation for future advancements for enhancing Polyglot LLMs performance in diverse linguistic contexts.

\newpage
\bibliography{neurips_2024}

\begin{thebibliography}{51}
\providecommand{\natexlab}[1]{#1}
\providecommand{\url}[1]{\texttt{#1}}
\expandafter\ifx\csname urlstyle\endcsname\relax
  \providecommand{\doi}[1]{doi: #1}\else
  \providecommand{\doi}{doi: \begingroup \urlstyle{rm}\Url}\fi

\bibitem[Abdin et~al.(2024)Abdin, Jacobs, Awan, Aneja, Awadallah, Awadalla, Bach, Bahree, Bakhtiari, Behl, et~al.]{abdin2024phi}
M.~Abdin, S.~A. Jacobs, A.~A. Awan, J.~Aneja, A.~Awadallah, H.~Awadalla, N.~Bach, A.~Bahree, A.~Bakhtiari, H.~Behl, et~al.
\newblock Phi-3 technical report: A highly capable language model locally on your phone.
\newblock \emph{arXiv preprint arXiv:2404.14219}, 2024.

\bibitem[Academy(2024)]{khanmigo}
K.~Academy.
\newblock Khanmigo.
\newblock Website, 2024.
\newblock URL \url{https://www.khanmigo.ai/}.
\newblock Accessed: 2024-05-21.

\bibitem[Ahuja et~al.(2023{\natexlab{a}})Ahuja, Hada, Ochieng, Jain, Diddee, Maina, Ganu, Segal, Axmed, Bali, et~al.]{mega}
K.~Ahuja, R.~Hada, M.~Ochieng, P.~Jain, H.~Diddee, S.~Maina, T.~Ganu, S.~Segal, M.~Axmed, K.~Bali, et~al.
\newblock Mega: Multilingual evaluation of generative ai.
\newblock \emph{arXiv preprint arXiv:2303.12528}, 2023{\natexlab{a}}.

\bibitem[Ahuja et~al.(2023{\natexlab{b}})Ahuja, Aggarwal, Gumma, Watts, Sathe, Ochieng, Hada, Jain, Axmed, Bali, et~al.]{ahuja2023megaverse}
S.~Ahuja, D.~Aggarwal, V.~Gumma, I.~Watts, A.~Sathe, M.~Ochieng, R.~Hada, P.~Jain, M.~Axmed, K.~Bali, et~al.
\newblock Megaverse: benchmarking large language models across languages, modalities, models and tasks.
\newblock \emph{arXiv preprint arXiv:2311.07463}, 2023{\natexlab{b}}.

\bibitem[AI(2023)]{anthropic2023claude3}
A.~AI.
\newblock Model card for claude 3, 2023.
\newblock URL \url{https://www-cdn.anthropic.com/de8ba9b01c9ab7cbabf5c33b80b7bbc618857627/Model_Card_Claude_3.pdf}.
\newblock Accessed: 2024-05-21.

\bibitem[AI(2024)]{sarvamai2024openhathi}
S.~AI.
\newblock Announcing openhathi series.
\newblock Sarvam AI Blog, 2024.
\newblock URL \url{https://www.sarvam.ai/blog/announcing-openhathi-series}.
\newblock Accessed: 2024-05-21.

\bibitem[AI4Bharat(2022)]{ai4bharat2022indicqa}
AI4Bharat.
\newblock Indicqa: A multilingual question answering dataset for 12 indic languages.
\newblock \url{https://huggingface.co/datasets/ai4bharat/IndicQA}, 2022.

\bibitem[Brown et~al.(2020)Brown, Mann, Ryder, Subbiah, Kaplan, Dhariwal, Neelakantan, Shyam, Sastry, Askell, Agarwal, Herbert{-}Voss, Krueger, Henighan, Child, Ramesh, Ziegler, Wu, Winter, Hesse, Chen, Sigler, Litwin, Gray, Chess, Clark, Berner, McCandlish, Radford, Sutskever, and Amodei]{DBLP:journals/corr/abs-2005-14165}
T.~B. Brown, B.~Mann, N.~Ryder, M.~Subbiah, J.~Kaplan, P.~Dhariwal, A.~Neelakantan, P.~Shyam, G.~Sastry, A.~Askell, S.~Agarwal, A.~Herbert{-}Voss, G.~Krueger, T.~Henighan, R.~Child, A.~Ramesh, D.~M. Ziegler, J.~Wu, C.~Winter, C.~Hesse, M.~Chen, E.~Sigler, M.~Litwin, S.~Gray, B.~Chess, J.~Clark, C.~Berner, S.~McCandlish, A.~Radford, I.~Sutskever, and D.~Amodei.
\newblock Language models are few-shot learners.
\newblock \emph{CoRR}, abs/2005.14165, 2020.
\newblock URL \url{https://arxiv.org/abs/2005.14165}.

\bibitem[Chen et~al.(2023)Chen, Pun, and Wang]{chen2023medprompt}
X.~Chen, C.-M. Pun, and S.~Wang.
\newblock Medprompt: Cross-modal prompting for multi-task medical image translation.
\newblock \emph{arXiv preprint arXiv:2310.02663}, 2023.

\bibitem[Clark et~al.(2020)Clark, Choi, Collins, Garrette, Kwiatkowski, Nikolaev, and Palomaki]{tydiqa}
J.~H. Clark, E.~Choi, M.~Collins, D.~Garrette, T.~Kwiatkowski, V.~Nikolaev, and J.~Palomaki.
\newblock Tydi qa: A benchmark for information-seeking question answering in typologically diverse languages.
\newblock \emph{Transactions of the Association for Computational Linguistics}, 2020.

\bibitem[Cohere(2024)]{cohere2024embedv3}
Cohere.
\newblock Introducing embed v3.
\newblock Cohere Blog, 2024.
\newblock URL \url{https://cohere.com/blog/introducing-embed-v3}.
\newblock Accessed: 2024-05-21.

\bibitem[corporation(2024{\natexlab{a}})]{azureopenai}
M.~corporation.
\newblock Azure openai service.
\newblock \url{https://azure.microsoft.com/en-us/products/cognitive-services/openai-service}, 2024{\natexlab{a}}.

\bibitem[corporation(2024{\natexlab{b}})]{azureopenaiturbo}
M.~corporation.
\newblock Gpt 3.5 turbo: Azure openai service.
\newblock \url{https://learn.microsoft.com/en-us/azure/cognitive-services/openai/concepts/models#chatgpt-gpt-35-turbo}, 2024{\natexlab{b}}.

\bibitem[corporation(2024{\natexlab{c}})]{azurepoenaigptfour}
M.~corporation.
\newblock Gpt-4: Azure openai service.
\newblock \url{https://learn.microsoft.com/en-us/azure/cognitive-services/openai/concepts/models#gpt-4-models}, 2024{\natexlab{c}}.

\bibitem[corporation(2024{\natexlab{d}})]{azuretranslate}
M.~corporation.
\newblock Azure cognitive services translator services.
\newblock \url{https://learn.microsoft.com/en-us/azure/cognitive-services/translator/}, 2024{\natexlab{d}}.
\newblock URL \url{https://learn.microsoft.com/en-us/azure/cognitive-services/translator/}.

\bibitem[Fernando et~al.(2023)Fernando, Banarse, Michalewski, Osindero, and Rockt{\"a}schel]{fernando2023promptbreeder}
C.~Fernando, D.~Banarse, H.~Michalewski, S.~Osindero, and T.~Rockt{\"a}schel.
\newblock Promptbreeder: Self-referential self-improvement via prompt evolution.
\newblock \emph{arXiv preprint arXiv:2309.16797}, 2023.

\bibitem[Gala et~al.(2024)Gala, Jayakumar, Husain, Khan, Kanojia, Puduppully, Khapra, Dabre, Murthy, Kunchukuttan, et~al.]{gala2024airavata}
J.~Gala, T.~Jayakumar, J.~A. Husain, M.~S. U.~R. Khan, D.~Kanojia, R.~Puduppully, M.~M. Khapra, R.~Dabre, R.~Murthy, A.~Kunchukuttan, et~al.
\newblock Airavata: Introducing hindi instruction-tuned llm.
\newblock \emph{arXiv preprint arXiv:2401.15006}, 2024.

\bibitem[Gao et~al.(2023)Gao, Xiong, Gao, Jia, Pan, Bi, Dai, Sun, and Wang]{gao2023retrieval}
Y.~Gao, Y.~Xiong, X.~Gao, K.~Jia, J.~Pan, Y.~Bi, Y.~Dai, J.~Sun, and H.~Wang.
\newblock Retrieval-augmented generation for large language models: A survey.
\newblock \emph{arXiv preprint arXiv:2312.10997}, 2023.

\bibitem[Goyal et~al.(2021)Goyal, Du, Ott, Anantharaman, and Conneau]{goyal2021larger}
N.~Goyal, J.~Du, M.~Ott, G.~Anantharaman, and A.~Conneau.
\newblock Larger-scale transformers for multilingual masked language modeling.
\newblock \emph{arXiv preprint arXiv:2105.00572}, 2021.

\bibitem[Green(2024)]{farmerchat}
D.~Green.
\newblock Farmer chat.
\newblock Website, 2024.
\newblock URL \url{https://farmerchat.digitalgreen.org/}.
\newblock Accessed: 2024-05-21.

\bibitem[He et~al.(2023)He, Lin, Gong, Jin, Zhang, Lin, Jiao, Yiu, Duan, Chen, et~al.]{he2023annollm}
X.~He, Z.~Lin, Y.~Gong, A.~Jin, H.~Zhang, C.~Lin, J.~Jiao, S.~M. Yiu, N.~Duan, W.~Chen, et~al.
\newblock Annollm: Making large language models to be better crowdsourced annotators.
\newblock \emph{arXiv preprint arXiv:2303.16854}, 2023.

\bibitem[Huang et~al.(2023)Huang, Tang, Zhang, Zhao, Song, Xia, and Wei]{huang2023not}
H.~Huang, T.~Tang, D.~Zhang, W.~X. Zhao, T.~Song, Y.~Xia, and F.~Wei.
\newblock Not all languages are created equal in llms: Improving multilingual capability by cross-lingual-thought prompting.
\newblock \emph{arXiv preprint arXiv:2305.07004}, 2023.

\bibitem[Hämmerl et~al.(2022)Hämmerl, Deiseroth, Schramowski, Libovický, Fraser, and Kersting]{hämmerl2022multilingual}
K.~Hämmerl, B.~Deiseroth, P.~Schramowski, J.~Libovický, A.~Fraser, and K.~Kersting.
\newblock Do multilingual language models capture differing moral norms?, 2022.

\bibitem[Joshi et~al.(2020)Joshi, Santy, Budhiraja, Bali, and Choudhury]{joshi-etal-2020-state}
P.~Joshi, S.~Santy, A.~Budhiraja, K.~Bali, and M.~Choudhury.
\newblock The state and fate of linguistic diversity and inclusion in the {NLP} world.
\newblock In \emph{Proceedings of the 58th Annual Meeting of the Association for Computational Linguistics}, pages 6282--6293, Online, July 2020. Association for Computational Linguistics.
\newblock \doi{10.18653/v1/2020.acl-main.560}.
\newblock URL \url{https://aclanthology.org/2020.acl-main.560}.

\bibitem[Khanuja et~al.(2021)Khanuja, Bansal, Mehtani, Khosla, Dey, Gopalan, Margam, Aggarwal, Nagipogu, Dave, et~al.]{khanuja2021muril}
S.~Khanuja, D.~Bansal, S.~Mehtani, S.~Khosla, A.~Dey, B.~Gopalan, D.~K. Margam, P.~Aggarwal, R.~T. Nagipogu, S.~Dave, et~al.
\newblock Muril: Multilingual representations for indian languages.
\newblock \emph{arXiv preprint arXiv:2103.10730}, 2021.

\bibitem[Kuchnik et~al.(2023)Kuchnik, Smith, and Amvrosiadis]{kuchnik2023validating}
M.~Kuchnik, V.~Smith, and G.~Amvrosiadis.
\newblock Validating large language models with relm.
\newblock \emph{Proceedings of Machine Learning and Systems}, 5, 2023.

\bibitem[Lewis et~al.(2019)Lewis, Oguz, Rinott, Riedel, and Schwenk]{DBLP:journals/corr/abs-1910-07475}
P.~S.~H. Lewis, B.~Oguz, R.~Rinott, S.~Riedel, and H.~Schwenk.
\newblock {MLQA:} evaluating cross-lingual extractive question answering.
\newblock \emph{CoRR}, abs/1910.07475, 2019.
\newblock URL \url{http://arxiv.org/abs/1910.07475}.

\bibitem[Littell et~al.(2017)Littell, Mortensen, Lin, Kairis, Turner, and Levin]{littell2017uriel}
P.~Littell, D.~R. Mortensen, K.~Lin, K.~Kairis, C.~Turner, and L.~Levin.
\newblock Uriel and lang2vec: Representing languages as typological, geographical, and phylogenetic vectors.
\newblock In \emph{Proceedings of the 15th Conference of the European Chapter of the Association for Computational Linguistics: Volume 2, Short Papers}, volume~2, pages 8--14, 2017.

\bibitem[Liu et~al.(2024{\natexlab{a}})Liu, Zhang, Zhao, Luu, and Bing]{liu2024translation}
C.~Liu, W.~Zhang, Y.~Zhao, A.~T. Luu, and L.~Bing.
\newblock Is translation all you need? a study on solving multilingual tasks with large language models.
\newblock \emph{arXiv preprint arXiv:2403.10258}, 2024{\natexlab{a}}.

\bibitem[Liu et~al.(2023)Liu, Li, Ji, and Lin]{liu2023towards}
X.~Liu, R.~Li, W.~Ji, and T.~Lin.
\newblock Towards robust multi-modal reasoning via model selection.
\newblock \emph{arXiv preprint arXiv:2310.08446}, 2023.

\bibitem[Liu et~al.(2024{\natexlab{b}})Liu, He, Han, Zhang, Liu, Tian, Zhang, Wang, Gao, Zhong, Pan, Xu, Wu, Liu, Zhang, Zhang, Hu, Zhang, Qiang, Liu, and Ge]{liu2024understanding}
Y.~Liu, H.~He, T.~Han, X.~Zhang, M.~Liu, J.~Tian, Y.~Zhang, J.~Wang, X.~Gao, T.~Zhong, Y.~Pan, S.~Xu, Z.~Wu, Z.~Liu, X.~Zhang, S.~Zhang, X.~Hu, T.~Zhang, N.~Qiang, T.~Liu, and B.~Ge.
\newblock Understanding llms: A comprehensive overview from training to inference, 2024{\natexlab{b}}.

\bibitem[Malaviya et~al.(2017)Malaviya, Neubig, and Littell]{malaviya17emnlp}
C.~Malaviya, G.~Neubig, and P.~Littell.
\newblock Learning language representations for typology prediction.
\newblock In \emph{Conference on Empirical Methods in Natural Language Processing (EMNLP)}, Copenhagen, Denmark, September 2017.

\bibitem[Microsoft(2023)]{m365copilot}
Microsoft.
\newblock Introducing microsoft 365 copilot – your copilot for work.
\newblock Microsoft Blog, 2023.
\newblock URL \url{https://blogs.microsoft.com/blog/2023/03/16/introducing-microsoft-365-copilot-your-copilot-for-work/}.
\newblock Accessed: 2024-05-21.

\bibitem[Ogundepo et~al.(2023)Ogundepo, Gwadabe, Rivera, Clark, Ruder, Adelani, Dossou, Diop, Sikasote, Hacheme, et~al.]{ogundepo2023afriqa}
O.~Ogundepo, T.~R. Gwadabe, C.~E. Rivera, J.~H. Clark, S.~Ruder, D.~I. Adelani, B.~F. Dossou, A.~A. Diop, C.~Sikasote, G.~Hacheme, et~al.
\newblock Afriqa: Cross-lingual open-retrieval question answering for african languages.
\newblock \emph{arXiv preprint arXiv:2305.06897}, 2023.

\bibitem[OpenAI(2023)]{openai2023gpt4}
OpenAI.
\newblock Gpt-4 technical report, 2023.

\bibitem[OpenAI(2024)]{openai2024ada3}
OpenAI.
\newblock New embedding models and api updates.
\newblock OpenAI Blog, 2024.
\newblock URL \url{https://openai.com/index/new-embedding-models-and-api-updates/}.
\newblock Accessed: 2024-05-21.

\bibitem[Ouyang et~al.(2022)Ouyang, Wu, Jiang, Almeida, Wainwright, Mishkin, Zhang, Agarwal, Slama, Ray, Schulman, Hilton, Kelton, Miller, Simens, Askell, Welinder, Christiano, Leike, and Lowe]{ouyang2022training}
L.~Ouyang, J.~Wu, X.~Jiang, D.~Almeida, C.~L. Wainwright, P.~Mishkin, C.~Zhang, S.~Agarwal, K.~Slama, A.~Ray, J.~Schulman, J.~Hilton, F.~Kelton, L.~Miller, M.~Simens, A.~Askell, P.~Welinder, P.~Christiano, J.~Leike, and R.~Lowe.
\newblock Training language models to follow instructions with human feedback, 2022.

\bibitem[Qin et~al.(2024)Qin, Chen, Zhou, Chen, Li, Liao, Li, Che, and Yu]{qin2024multilingual}
L.~Qin, Q.~Chen, Y.~Zhou, Z.~Chen, Y.~Li, L.~Liao, M.~Li, W.~Che, and P.~S. Yu.
\newblock Multilingual large language model: A survey of resources, taxonomy and frontiers, 2024.

\bibitem[Rajpurkar et~al.(2016)Rajpurkar, Zhang, Lopyrev, and Liang]{DBLP:journals/corr/RajpurkarZLL16}
P.~Rajpurkar, J.~Zhang, K.~Lopyrev, and P.~Liang.
\newblock Squad: 100, 000+ questions for machine comprehension of text.
\newblock \emph{CoRR}, abs/1606.05250, 2016.
\newblock URL \url{http://arxiv.org/abs/1606.05250}.

\bibitem[Research(2024)]{shiksha}
M.~Research.
\newblock Teachers in india help microsoft research design ai tool for creating great classroom content.
\newblock Microsoft Research Blog, 2024.
\newblock Accessed: 2024-05-21.

\bibitem[Sahoo et~al.(2024)Sahoo, Singh, Saha, Jain, Mondal, and Chadha]{sahoo2024systematic}
P.~Sahoo, A.~K. Singh, S.~Saha, V.~Jain, S.~Mondal, and A.~Chadha.
\newblock A systematic survey of prompt engineering in large language models: Techniques and applications.
\newblock \emph{arXiv preprint arXiv:2402.07927}, 2024.

\bibitem[Shen et~al.(2024)Shen, Song, Tan, Li, Lu, and Zhuang]{shen2024hugginggpt}
Y.~Shen, K.~Song, X.~Tan, D.~Li, W.~Lu, and Y.~Zhuang.
\newblock Hugginggpt: Solving ai tasks with chatgpt and its friends in hugging face.
\newblock \emph{Advances in Neural Information Processing Systems}, 36, 2024.

\bibitem[Shi et~al.(2022)Shi, Suzgun, Freitag, Wang, Srivats, Vosoughi, Chung, Tay, Ruder, Zhou, et~al.]{shi2022language}
F.~Shi, M.~Suzgun, M.~Freitag, X.~Wang, S.~Srivats, S.~Vosoughi, H.~W. Chung, Y.~Tay, S.~Ruder, D.~Zhou, et~al.
\newblock Language models are multilingual chain-of-thought reasoners.
\newblock \emph{arXiv preprint arXiv:2210.03057}, 2022.

\bibitem[Sitaram et~al.(2023)Sitaram, Choudhury, Patra, Chaudhary, Ahuja, and Bali]{sitaram-etal-2023-everything}
S.~Sitaram, M.~Choudhury, B.~Patra, V.~Chaudhary, K.~Ahuja, and K.~Bali.
\newblock Everything you need to know about multilingual {LLM}s: Towards fair, performant and reliable models for languages of the world.
\newblock In Y.-N.~V. Chen, M.~Margot, and S.~Reddy, editors, \emph{Proceedings of the 61st Annual Meeting of the Association for Computational Linguistics (Volume 6: Tutorial Abstracts)}, pages 21--26, Toronto, Canada, July 2023. Association for Computational Linguistics.
\newblock \doi{10.18653/v1/2023.acl-tutorials.3}.
\newblock URL \url{https://aclanthology.org/2023.acl-tutorials.3}.

\bibitem[Team et~al.(2023)Team, Anil, Borgeaud, Wu, Alayrac, Yu, Soricut, Schalkwyk, Dai, Hauth, et~al.]{team2023gemini}
G.~Team, R.~Anil, S.~Borgeaud, Y.~Wu, J.-B. Alayrac, J.~Yu, R.~Soricut, J.~Schalkwyk, A.~M. Dai, A.~Hauth, et~al.
\newblock Gemini: a family of highly capable multimodal models.
\newblock \emph{arXiv preprint arXiv:2312.11805}, 2023.

\bibitem[Touvron et~al.(2023)Touvron, Martin, Stone, Albert, Almahairi, Babaei, Bashlykov, Batra, Bhargava, Bhosale, et~al.]{touvron2023llama}
H.~Touvron, L.~Martin, K.~Stone, P.~Albert, A.~Almahairi, Y.~Babaei, N.~Bashlykov, S.~Batra, P.~Bhargava, S.~Bhosale, et~al.
\newblock Llama 2: Open foundation and fine-tuned chat models.
\newblock \emph{arXiv preprint arXiv:2307.09288}, 2023.

\bibitem[Vizcaíno et~al.(2021)Vizcaíno, Saltarin, Belyaev, Lyck, Lasser, and Favaro]{vizcaino2021convnd}
J.~P. Vizcaíno, F.~Saltarin, Y.~Belyaev, R.~Lyck, T.~Lasser, and P.~Favaro.
\newblock Learning to reconstruct confocal microscopy stacks from single light field images.
\newblock \emph{IEEE Transactions on Computational Imaging}, 7:\penalty0 775--788, 2021.
\newblock \doi{10.1109/TCI.2021.3097611}.

\bibitem[Wang et~al.(2020)Wang, Tsvetkov, and Neubig]{DBLP:journals/corr/abs-2004-06748}
X.~Wang, Y.~Tsvetkov, and G.~Neubig.
\newblock Balancing training for multilingual neural machine translation.
\newblock \emph{CoRR}, abs/2004.06748, 2020.
\newblock URL \url{https://arxiv.org/abs/2004.06748}.

\bibitem[Wei et~al.(2022)Wei, Wang, Schuurmans, Bosma, Xia, Chi, Le, Zhou, et~al.]{weichain}
J.~Wei, X.~Wang, D.~Schuurmans, M.~Bosma, F.~Xia, E.~H. Chi, Q.~V. Le, D.~Zhou, et~al.
\newblock Chain-of-thought prompting elicits reasoning in large language models.
\newblock In \emph{Advances in Neural Information Processing Systems}, 2022.

\bibitem[Yang et~al.(2022)Yang, Lin, Yang, Wang, Zhou, and Yang]{yang2022prompt}
H.~Yang, J.~Lin, A.~Yang, P.~Wang, C.~Zhou, and H.~Yang.
\newblock Prompt tuning for generative multimodal pretrained models.
\newblock \emph{arXiv preprint arXiv:2208.02532}, 2022.

\bibitem[Zohar et~al.(2024)Zohar, Huang, Wang, and Yeung]{zohar2024lovm}
O.~Zohar, S.-C. Huang, K.-C. Wang, and S.~Yeung.
\newblock Lovm: Language-only vision model selection.
\newblock \emph{Advances in Neural Information Processing Systems}, 36, 2024.

\end{thebibliography}
\bibliographystyle{abbrvnat}
\section*{Appendix}

\section{Similar Language Algorithm}
\label{app:sim}
Section 3 introduced various prompt strategies and prompt templates that we have optimized for polyglot LLMs. One of the prompt strategies defined is round-tripping the input in source language through "Similar high-resourced language (Sim)". In this section, we present the algorithm for identifying the right set of similar high-resourced languages for a given source language. For every language, we associate its class attribute between 0-5 based on the classes defined in \cite{joshi-etal-2020-state}. Here, class 5 represents very high-resourced languages like English, whereas 0 represents very low-resourced languages like Gondi, Mundari, etc. We use the language similarity metrics based on  language feature similarities\cite{malaviya17emnlp} captured in lang2vec \cite{littell2017uriel}. We give higher preference to the languages with Latin script since the languages with Latin script have shown better performance on GPTx models\cite{mega}.

\begin{algorithm}
\caption{Get language relevance score based on language similarity distance, the class of related language and whether the related language has Latin script.}
\label{alg:lang_relevance_score}
\SetKwFunction{FMain}{GetRelevanceScore}
\SetKwProg{Fn}{Function}{:}{}
$w_{Latin} \gets 0.9$;

\Fn{\FMain{$d$, $l_{cls}$, $isLatin$}}
  {
  $w \gets 1$; 
  
  \If{$isLatin$}
  {$w \gets w_{Latin}$}
  $score \gets w \times d/l_{cls}$;
  
  return $score$
  }  
\end{algorithm}

\begin{algorithm}
\caption{Identifying similar high-resourced languages for a given language}\label{alg:similar_language}
\KwData{Source language $l_{s}$}
\KwResult{A set of similar high-resourced languages $L_{similar}$}
$L_{similar} \gets \emptyset$ \\
Language class threshold $cls_{threshold} \gets 3$ \\
Languages similarity distance threshold $dist_{threshold} \gets 0.5$ \\
\For {$ l \in L$}
{
    \If {$class(l) \geq cls_{threshold}$}
    {
Similarity distance between $l$ and $l_S$: $d \gets lang2vec\_distance(['syntactic','genetic','geographic'],l,l_{s})$;
$RelevanceScore \gets GetRelevanceScore(average(d), class(l), isLatin(l)$;

\If{$RelevanceScore \leq dist_{threshold}$}
{
$L_{similar}.add(l)$
}
}
}
\end{algorithm}

\section{Prompt Strategies Results}
\label{app:prompt}
\textbf{Performance of prompts for \tydi.}

\begin{table}
\centering
\resizebox{0.5\textwidth}{!}{%
\begin{tabular}{lll|ll}
\hline
     & \multicolumn{2}{l|}{\meta}    & \multicolumn{2}{l}{\gptevalf} \\ \hline
     & \gptt & \turbo & \gptt & \turbo \\ \hline
Mono & \textbf{0.64}        & \textbf{0.64}         & \textbf{0.71}        & \textbf{0.71}         \\
Tans     & 0.49 & 0.51 & 0.61 & 0.63 \\
simi     & 0.47 & 0.47 & 0.58 & 0.58 \\
Aggsrc   & 0.62 & 0.63 & 0.69 & 0.70 \\
aggtrans & 0.49 & 0.52 & 0.60 & 0.63 \\ \hline
\end{tabular}%
}
\caption{Performance of different Prompt strategies for \tydi}
\label{tab:prompt-tydi}
\end{table}

In this section we present the performance of our Prompts on \tydi dataset, We report \meta and \gptevalf for each prompt Averaged across all 9 languages. The numbers are reported for \gptt and \turbo with text-embedding-ada-002 embeddings In Table.\ref{tab:prompt-tydi} we observe similar trends to experiment with \indic, i.e., Each model have different trend across the different prompt strategy and the choice of the metrics also favours different model making the it difficult to find a suitable choice of prompt for a generalized pipeline.

\textbf{Per language performance for \gptt and \turbo for \indic}

Table. \ref{tab:gpt4-indic-prompt}, \ref{tab:turbo-indic-prompt} presents the performance of \gptt and \turbo respectively with text-embedding-ada-002 embeddings, across all 11 languages and 5 prompts that we propose. Here we observe strong patterns for Agg\_Sim performing the best across majority of the languages ( \(\frac{7}{11}\) for \gptt and \(\frac{5}{11}\) for \turbo), Mono performs better and comes very close to Agg\_sim in these languages. For languages such as "ta", "te" translate is prefered. With the limited languages the variance in the trend is high and a rule based system would fail with inclusion of more languages.  
\begin{table}
\begin{minipage}[t]{0.5\linewidth}
    \centering
    \caption{\gptt on \indic}
    \label{tab:gpt4-indic-prompt}
    \resizebox{1.0\columnwidth}{!}{%
    \begin{tabular}{llllll}
    Lang & Mono          & Translate     & Similar & AggSim                            & AggTrans      \\ \hline
    as   & \textbf{0.58} & 0.33          & 0.33    & \textbf{0.58}                     & 0.32          \\
    bn   & \textbf{0.62} & 0.38          & 0.36    & \textbf{0.62}                     & 0.36          \\
    gu   & \textbf{0.59} & 0.31          & 0.30    & \textbf{0.59}                     & 0.30          \\
    hi   & \textbf{0.67} & 0.54          & 0.42    & {\textbf{0.68}} & 0.51          \\
    kn   & \textbf{0.48} & 0.31          & 0.25    & \textbf{0.48}                     & 0.29          \\
    ml   & \textbf{0.32} & 0.30          & 0.19    & \textbf{0.32}                     & 0.29          \\
    mr   & \textbf{0.58} & 0.33          & 0.30    & \textbf{0.57}                     & 0.32          \\
    or   & \textbf{0.57} & 0.29          & 0.27    & \textbf{0.57}                     & 0.27          \\
    pa   & \textbf{0.61} & 0.46          & 0.43    & 0.60                              & 0.46          \\
    ta   & \textbf{0.31} & \textbf{0.40} & null    & 0.34                              & 0.39          \\
    te   & 0.25          & 0.36          & 0.17    & 0.28                              & \textbf{0.37} \\ \hline
    AVG  & \textbf{0.51} & 0.36          & 0.30    & \textbf{0.51}                     & 0.35          \\ \hline
    \end{tabular}%
    }
    
\end{minipage}
~
\begin{minipage}[t]{0.5\linewidth}
    \caption{\turbo on \indic}
    \label{tab:turbo-indic-prompt}
    \centering
    \resizebox{1.0\columnwidth}{!}{%
    \begin{tabular}{lrrlll}
    Lang & {Mono}          & {Translate}     & Similar                  & AggSim                            & AggTrans                 \\ \hline
    as   & {\textbf{0.40}} & {0.34}          & 0.33                     & \textbf{0.45}                     & 0.37                     \\
    bn   & {\textbf{0.54}} & {0.39}          & {0.34} & \textbf{0.54}                     & 0.46                     \\
    gu   & \textbf{0.48}                     & 0.32                              & {0.30} & {\textbf{0.49}} & {0.33} \\
    hi   & \textbf{0.63}                     & 0.52                              & 0.38                     & \textbf{0.64}                     & 0.52                     \\
    kn   & \textbf{0.47}                     & 0.32                              & {0.21} & \textbf{0.46}                     & {0.31} \\
    ml   & \textbf{0.23}                     & \textbf{0.32}                     & 0.13                     & \textbf{0.26}                     & 0.31                     \\
    mr   & \textbf{0.48}                     & 0.34                              & 0.30                     & \textbf{0.47}                     & 0.36                     \\
    or   & \textbf{0.40}                     & 0.29                              & {0.27} & \textbf{0.39}                     & 0.32                     \\
    pa   & \textbf{0.54}                     & 0.46                              & {0.40} & \textbf{0.54}                     & 0.44                     \\
    ta   & {\textbf{0.31}} & {\textbf{0.40}} & null                     & 0.24                              & \textbf{0.39}            \\
    te   & {0.25}          & {\textbf{0.36}} & 0.17                     & 0.25                              & \textbf{0.34}            \\ \hline
    AVG  & {\textbf{0.43}} & {0.37}          & 0.28                     & \textbf{0.43}                     & 0.38                     \\ \hline
    \end{tabular}%
    }
    
\end{minipage}
\end{table}


\textbf{Per language performance for \gptt and \turbo for \tydi}
In Table. \ref{tab:gptt-tydi-prompt}, \ref{tab:turbo-tydi-prompt} performance of \gptt and \turbo along with text-embedding-ada-002 embeddings are presented across all 9 languages and 5 proposed prompts. Here contrary to \indic experiments Mono is preferred over Agg\_sim making a significant change in distribution. The optimal prompt doesn't depend only on the language or model but also on the distribution of the question, this statement is supported by the fact \tydi and \indic share 2 languages "bn" and "te", while in \indic Agg\_sim was prefered for "bn" and Translate for "te" it has completely shifted to Mono for both "bn" and "te" in \tydi. Hence prompt selection is depends on the language and also the distribution of the dataset or sample.

\begin{table}
\begin{minipage}[t]{0.5\linewidth}
    \centering
    \caption{\gptt on \tydi}
    \label{tab:gptt-tydi-prompt}
    \resizebox{1.0\columnwidth}{!}{%
    \begin{tabular}{llllll}
    Lang & Mono          & Translate     & Similar & AggSim        & AggTrans \\ \hline
    ar   & \textbf{0.50} & 0.43          & null    & \textbf{0.50} & 0.40     \\
    bn   & \textbf{0.69} & 0.46          & 0.43    & \textbf{0.69} & 0.43     \\
    en   & \textbf{0.65} & null          & 0.60    & 0.62 & 0.58     \\
    fi   & \textbf{0.63} & 0.49          & 0.48    & 0.59 & 0.49     \\
    id   & \textbf{0.66} & 0.58          & 0.53    & 0.63 & 0.54     \\
    ko   & \textbf{0.64} & {0.48}        & 0.43    & {0.63} & 0.47     \\
    ru   & \textbf{0.51} & 0.45          & 0.46    & {0.50} & 0.44     \\
    sw   & \textbf{0.80} & 0.63          & null    & {0.78} & 0.65     \\
    te   & \textbf{0.67} & 0.42          & 0.39    & {0.66} & 0.43     \\ \hline
    AVG  & \textbf{0.64} & 0.49          & 0.47    & {0.62} & 0.49     \\ \hline
    \end{tabular}%
    }
\end{minipage}
~
\begin{minipage}[t]{0.5\linewidth}
    \centering
    \caption{\turbo on \tydi}
    \label{tab:turbo-tydi-prompt}
    \resizebox{1.0\columnwidth}{!}{%
    \begin{tabular}{llllll}
    Lang & Mono          & Translate & Similar & AggSim & AggTrans \\ \hline
    ar   & 0.53          & 0.45      & null    & 0.52   & 0.42     \\
    bn   & 0.65          & 0.46      & 0.41    & 0.65   & 0.48     \\
    en   & 0.66          & null      & 0.61    & 0.65   & 0.64     \\
    fi   & 0.68          & 0.53      & 0.50    & 0.65   & 0.54     \\
    id   & 0.67          & 0.61      & 0.53    & 0.66   & 0.59     \\
    ko   & 0.65          & 0.49      & 0.46    & 0.66   & 0.51     \\
    ru   & 0.52          & 0.46      & 0.45    & 0.51   & 0.44     \\
    sw   & 0.76          & 0.64      & null    & 0.74   & 0.64     \\
    te   & 0.66          & 0.44      & 0.36    & 0.67   & 0.43     \\ \hline
    AVG  & \textbf{0.64} & 0.51      & 0.47    & 0.63   & 0.52     \\ \hline
    \end{tabular}%
    }
\end{minipage}
\end{table}


\section{Hybrid approach}
\label{app:hybrid}

In this section we evaluate the performance of our Hybrid Approach across text-ada-002-embedding, Adav3, XLMRXXL and Cohere embed\_multilingual\_v3. We use \tydi as the dataset and average the \meta and \gptevalf across all 9 languages and all 5 prompts. In Table. \ref{tab:hybrid-perf-tydi} we present the values for both \gptt and \turbo, while the trend is completely different to that of \indic which could be primarily attributed to the languages typology and derivations.

\begin{table}
\centering
\resizebox{0.7\textwidth}{!}{%
\begin{tabular}{llllll}
\hline
Metrics & Models                & Ada           & Adav3 & XLMRXXL & Cohere        \\ \hline
\multirow{2}{*}{\meta}     & \gptt & \textbf{0.64} & \textbf{0.64} & 0.60 & \textbf{0.61} \\
        & \turbo & \textbf{0.64} & 0.60  & 0.57    & \textbf{0.59} \\ \hline
\multirow{2}{*}{\gptevalf} & \gptt & \textbf{0.71} & \textbf{0.71} & 0.65 & \textbf{0.68} \\
        & \turbo & \textbf{0.71} & 0.66  & 0.63    & \textbf{0.66} \\ \hline
\end{tabular}%
}
\caption{Hybrid approach performance - \tydi.}
\label{tab:hybrid-perf-tydi}
\end{table}







    









\section{Detailed Training Procedure \& Implemenation Details}\label{app:training}\label{app:hybrid_arch}

The algorithm employs separate strategies for inference and training tailored to different operational conditions. During inference, the algorithm selects the optimal configuration based on F1 score predictions from task and configuration embeddings as described in Algorithm \ref{alg:learning}. In the Offline Setting, configuration selection is deterministic, using an argmax function for precise, data-rich environments. Conversely, the Online Setting uses a probabilistic softmax function to adapt to data-scarce situations, enabling dynamic exploration and refinement of configurations.

For training, the offline mode applies a Mean Squared Error (MSE) loss across all configurations, ensuring comprehensive learning. In contrast, the online mode implements a sparse MSE loss, updating only the evaluated configurations through a masking technique. This approach reduces computational load and accelerates adaptation to new data, optimizing performance in real-time applications as outlined in Algorithm \ref{alg:learning}.

The sparse MSE Loss employs a Mask \(\mathcal{M}\) which is defined as,
Let \(\mathbf{C}\) be a tensor of order \(m\) with dimensions \(n_1 \times n_2 \times \dots \times n_m\). Suppose \(\hat{c} = C_{i_1, i_2, \dots, i_m}\) is a selected element from \(\mathbf{C}\), where \((i_1, i_2, \dots, i_m)\) are the indices of \(\hat{c}\) in \(\mathbf{C}\). Define the tensor \(\mathcal{M}\) as follows:

\begin{equation}
\mathcal{M}_{j_1, j_2, \dots, j_m} = 
\begin{cases} 
1 & \text{if } (j_1, j_2, \dots, j_m) = (i_1, i_2, \dots, i_m) \\
0 & \text{otherwise}
\end{cases}
\label{eq:mask_matrix}
\end{equation}

\begin{algorithm}[H]
\caption{Learning Strategy Algorithm for Inference and Training}\label{alg:learning}
\DontPrintSemicolon  

\KwData{Task descriptions \(\mathcal{T}\), configuration options \(\mathcal{C}_i\)}
\KwResult{Optimal configuration \(\hat{c}\) and its corresponding F1 score}

\(\mathcal{B}\) - LLaMa-2-70B backbone for embedding generation\;
\(\mathcal{H}\) - Conv-ND layers for F1 score prediction\;
\(e\) - embedding projection size, \(e = 8192\)\;
\(m\) - number of parameters, \(m = 3\) (e.g., language model, embedding model, prompt strategies)\;
\(\mathcal{R}^{n_1 \times n_2 \times \dots \times n_m}\) - size of the N-dimensional array for configurations\;
\(bs\) - Batch size of Task Definitions.

\For{\(\mathcal{T}_{j} \leftarrow \{\mathcal{T}_{0},...\mathcal{T}_{bs} \} \)}{
    \;
    
    \(\mathcal{E}_{Tj} \leftarrow \mathcal{B}(\mathcal{T}_{j})\) ;
    \(\mathcal{E}_{Ci} \leftarrow \mathcal{B}(\mathcal{C}_{i})\) \\
    \(\mathcal{E}_{j} \leftarrow \text{Concatenate}(\mathcal{E}_{Tj}, \mathcal{E}_{Ci})\) 
    
    \(\hat{y} \leftarrow \mathcal{H}(\mathcal{E}_{j})\) 
    
    \;
    
    {\color{blue}\tcc{Inference for selecting configuration}}

    \uIf{Offline Setting}{
    \(\hat{c} \leftarrow \arg\max(\hat{y})\)
    }\uElseIf{Online Setting}{
    \(\hat{c} \sim \text{Softmax}(\hat{y})\)
    }
    
    \;
    
    {\color{blue}\tcc{Training to update \(\mathcal{H}\) \& \(\mathcal{B}\) }}

    \uIf{Offline Setting}{
        \(y \leftarrow \text{Ground truth F1 scores }  \forall \mathcal{C}_{i} \)\;
        \(Loss_{\text{off}} \leftarrow \text{MSE}(\hat{y}, y)\)
    }
    \uElseIf{Online Setting}{
        \(y_{\text{sparse}} \leftarrow \text{Ground truth F1 score for } \hat{c} \)\;
        \(\mathcal{M} \leftarrow \text{Mask matrix using eq. \ref{eq:mask_matrix} }\)\;
        \(Loss_{\text{on}} \leftarrow \text{MSE}(\mathcal{M} \odot \hat{y}, y_{\text{sparse}})\)
    }

    \;
    
    Update \(\mathcal{H}\) \& \(\mathcal{B}\) using \(Loss\)
}
\end{algorithm}

\subsection{Implementation Details}
\label{sec:imple}
In this work, we use Azure OpenAI models~\cite{azureopenai} for all our LLM and embedding models including \gptt, \turbo and \mix. Furthermore, for the learnign model, we train the Llama model on GPU with 4 x A100 80 GB, CPU	with 96 cpu cores at 2.2GHz and 1024 GB RAM. The duration of training 100 offline epochs is 1.42 Hrs. The duration of training 25 online epochs is 0.74 Hrs. The inference and evaluation is dependent on the rate limits imposed by Azure OpenAI APIs~\cite{azureopenai}.
\textbf{Model Version}: For LLMs we use \gptt - 0125-preview, \turbo - 0125, Mixtral - Mixtral-8x7B-Instruct-v0.1; For Embeddings we use ada - text-ada-002-embedding, ada3 - text-ada-003-embedding, XLMR-XXL - facebook/xlm-roberta-xxl and Cohere - embed\_multilingual\_v3; 

\section{\gpteval Setup and details}
\label{sec:gptanno}

\subsection{Human Annotation Task Details}
\label{subsec:human_annotation}

We build a simple human annotation interface using Streamlit\footnote{https://streamlit.io/} where the context, the question related to the context, and the ground truth answer for each record are fetched from the IndicQA dataset\cite{ai4bharat2022indicqa}. In this evaluation task, the annotators are first presented with a passage that acts as the context required to answer the question which is shown along with the ground truth answer. The annotators are then asked to evaluate the answers generated by the LLM using different strategies based on the ground truth answer provided, by answering one of the following options: "Yes", "No" or "Partial".  Here is the instruction provided to the annotators. 

\fbox{\begin{minipage}{40em}
First, select your language and go through the context under the title "Context GT" once. Then, look at the question and try to answer this question and compare it with the ground truth answer. Next, for all the available answers, choose:
\begin{enumerate}
    \item "Yes" if the answer is absolutely correct(minor punctuation errors are allowed)
    \item "Partial" if the answer captures some part of the core answer, but has grammatical mistakes or minor errors(spelling, etc.) that make the answer partially correct.
    \item "No" if the answer is completely wrong
\end{enumerate}
\end{minipage}}

Based on the human annotations for each question, we then recompute the F1 score. The updated F1 scores are calculated using Algorithm \ref{alg:eval}, where $evals$ contains evaluations for all the strategies annotated by the human annotator.


\begin{algorithm}
\caption{Evaluation Algorithm when using Human Annotator or \gpteval}\label{alg:eval}
\KwData{$ground\_truth$, $gpt\_answers$, $evals$}
\KwResult{$eval\_scores$}
$eval\_scores \gets []$;
$valid\_answers \gets []$\;
$evals = get\_eval(gpt\_answers)$;
$valid\_answers.append(ground\_truth)$;
\For{$i\gets0$ \KwTo $len(gpt\_answers)$} {
    \If{$evals[i] = "Yes"$} {
    $valid\_answers.append(gpt\_answers[i])$;
    }
}
\For{$i\gets0$ \KwTo $len(gpt\_answers)$} {
    $eval\_f1.append(compute\_score(gpt\_answers[i], valid\_answers))$;
}
\end{algorithm}

\subsection{GPT Eval process}

In Section \ref{sec:limitations}, we introduced \gpteval, where GPT models perform the evaluation of the answer generated when compared to the ground truth. Similar to the human evaluation task described in the previous subsection \ref{subsec:human_annotation}, the \gpteval is tasked to evaluate the LLM responses based on the available ground truth for the given record. The prompt below is used for \turbo in order to evaluate the answers.

\fbox{
\begin{minipage}{40em}
You are a multilingual evaluation assistant. Users will send in a query, context text, the correct answer for the query based on the context text, and also an answer that needs to be evaluated. You will evaluate the answer based on the context text and the correct answer that the user has sent and respond with Yes, No, or Partial based on the below evaluation instructions. Instructions: 1. Yes if the answer is absolutely correct. 2. Partial if the answer captures some part of the correct answer, but has minor errors like grammatical or spelling mistakes, etc. 3.No if the answer is completely wrong.
\end{minipage}}

The updated F1 Scores for each of the strategy is calculated using Algorithm \ref{alg:eval} where $evals$ contains "Yes", "No" or "Partial" evaluations as judged by the \gpteval.
\end{document}